# Hybrid Deep Learning Framework for Enhanced Diabetic Retinopathy Detection: Integrating Traditional Features with AI-driven Insights

Arpan Maity, Aviroop Pal, MD. Samiul Islam, and Tamal Ghosh

Department of Computer Science and Engineering, Adamas University, Barbaria, Barasat 700126, India
`tamal.ghosh1@adamasuniversity.ac.in`

**Abstract.** Diabetic Retinopathy (DR), a vision-threatening complication of Diabetes Mellitus (DM), is a major global concern, particularly in India, which has one of the highest diabetic populations. Prolonged hyperglycemia damages retinal microvasculature, leading to DR symptoms like microaneurysms, hemorrhages, and fluid leakage, which, if undetected, cause irreversible vision loss. Therefore, early screening is crucial as DR is asymptomatic in its initial stages. Fundus imaging aids precise diagnosis by detecting subtle retinal lesions. This paper introduces a hybrid diagnostic framework combining traditional feature extraction and deep learning (DL) to enhance DR detection. While handcrafted features capture key clinical markers, DL automates hierarchical pattern recognition, improving early diagnosis. The model synergizes interpretable clinical data with learned features, surpassing standalone DL approaches that demonstrate superior classification and reduce false negatives. This multimodal AI-driven approach enables scalable, accurate DR screening, crucial for diabetes-burdened regions.

## 1 Introduction

Diabetes is a widespread epidemic, particularly in India, contributing to conditions like Diabetic Retinopathy (DR). Developing automated diagnostic models is crucial to support ophthalmologists and reduce patient morbidity. Globally, diabetes affects 422 million people, with India ranking among the top three countries with the highest diabetic population [1]. The number of diabetics has surged from 108 million to 422 million, with half residing in India, China, the USA, Brazil, and Indonesia. According to a Lancet study, China, India, and the USA have the highest diabetic populations. Diabetes can lead to numerous complications, including damage to the optic nerve in advanced stages. DR is caused by damage to retinal blood vessels due to diabetes, leading to irreversible vision loss. Early retinal screening can help diagnose retinal damage at the initial stages. DR is often asymptomatic initially, leaving many patients unaware until their vision is affected. Therefore, early and regular screening is essential to prevent further complications and control disease progression. A key symptom of DR is the presence of exudates, which can be detected in fundus images of the eyes, indicating the development or presence of DR. Identifying lesions in fundus images can also aid in early detection [2]. Ophthalmologists study color fundus images to identify features related to DR, such as hemorrhages, soft and hard exudates, and microaneurysms (MA), which appear as tiny red dots due to local distension of capillary walls. DR progresses through four stages: Mild non-proliferative retinopathy is characterized by the presence of microaneurysms, marking the earliest stage. Moderate non-proliferative retinopathy occurs when blood vessels nourishing the retina may distort and swell, losing their ability to transport blood. Severe non-proliferative retinopathy results from the blockage of more blood vessels, depriving the retina of blood supply and signaling the growth of new blood vessels. Proliferative diabetic retinopathy (PDR) is the advanced stage where new blood vessels proliferate along the retina and into the vitreous gel. These fragile vessels bleed and leak frequently, and associated scar tissue may cause retinal detachment, leading to permanent vision loss [3].

To aid this, we proposed developing an optimal deep net model by integrating traditional feature extraction with DL techniques. This combined approach aims to surpass the performance of existing models such as ResNet50, VGG-16, VGG-19, ResNet101, MobileNetV2, MobileNet, InceptionV3, InceptionResNetV2, DenseNet169, DenseNet121, and XceptionNet. The focus is on utilizing deep learning feature extraction architectures to improve the accuracy and efficiency of diagnosing Diabetic Retinopathy.

## 2    Related Work

Diabetic Retinopathy (DR) is a diabetes complication with the possibility of leading to vision impairment and blindness if detected and treated late. There has been considerable research work involved in developing automated detection and classification methods of DR over the years through machine learning (ML), image processing, and artificial intelligence.

The ML-based object detection system is an improved and precise diagnostic approach for the automated retinal abnormalities detection process [1]. Microaneurysms-focused image analysis-based early disease detection can help avoid disease progression [2]. Enhancement of the quality of retinal images using superior contrast treatment could facilitate better detection of DR [3]. For diabetic and malarial retinopathy the images are analyzed using saliency estimation method based on intensity and compactness for leakage detection [4]. ef. [5] emphasized early DR detection via digital fundus imaging using segmentation and classification. Ref. [6] developed a system for detecting retinal lesions like hemorrhages and microaneurysms, crucial for assessing DR severity. Ref. [7] compared various image processing techniques for segmentation and classification, while Ref. [8] explored automatic retinal feature extraction applicable to DR screening. Ref. [9] pioneered the use of Artificial Neural Networks (ANNs) for DR detection. Refs. [10-12] analyzed computational and ML-based approaches for automated DR grading, enhancing diagnosis. Ref. [13] evaluated imaging transforms, confirming their role in improving retinal structure visualization. Ref. [14] introduced texture-based analysis for DR classification, and Ref. [15] proposed the Xception model, a DL-based feature extractor demonstrating efficiency in DR detection.

Improvements in ML, DL, and image processing have greatly enhanced DR detection and classification with improved accuracy. With advancements in artificial intelligence-based diagnostic technology, it is increasingly becoming a valuable resource in assisting ophthalmologists in early detection and treatment planning.

## 3    Methodology

The dataset used in this study is based on diabetic retinopathy (size of 224×224), consisting of retina scan images for detecting diabetic retinopathy. This dataset, freely available on *https://www.kaggle.com/datasets/sovitrath/diabetic-retinopathy-224x224-2019-data* includes a total of 3,622 images for training and testing. The images are categorized into five classes based on the severity of diabetic retinopathy: No_DR, Mild, Moderate, Severe, and Proliferative_DR as shown in Figure 1.

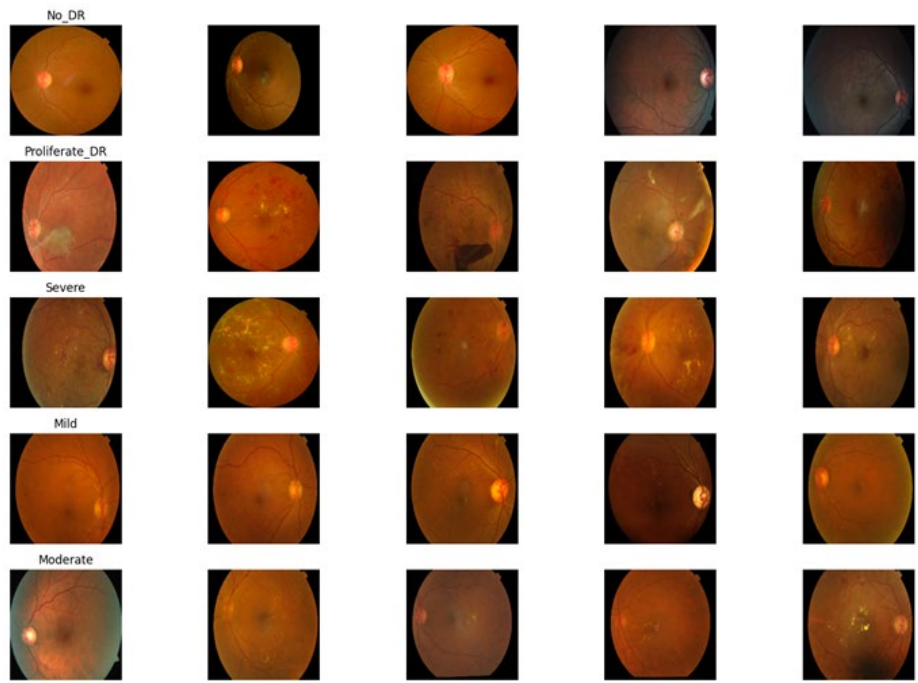

**Fig. 1.** Categorization of images based on classes

### 3.1 Image Segmentation

Segmentation is a crucial step in image processing, used to identify the region of interest (ROI) while eliminating unnecessary noise that does not contribute to feature extraction. The test dataset comprises various types of diabetic retinopathy images, collected from a diverse set of retina scans captured through fundus photography under different imaging conditions. Using segmentation, the ROI is effectively highlighted, and the corresponding ROI mask images are saved to enhance feature extraction for improved analysis. Segmentation techniques are as follows,

**Gaussian Blur:** Gaussian blur [16] is a widely used image processing technique that applies a convolution operation to smooth an image, reducing noise and detail. It uses a Gaussian function to create a kernel, which is then convolved with the image..

**Adaptive Thresholding:** Adaptive thresholding is an image segmentation technique that dynamically determines thresholds for small regions of an image, making it suitable for images with varying illumination [17]. Unlike global thresholding, it computes a unique threshold $T(x,y)$ for each pixel based on its local neighborhood.

**Morphological Opening:** Morphological opening is a fundamental operation in mathematical morphology used to remove small objects or noise from an image while preserving the overall shape of larger structures [18].

### 3.2 Traditional featured extrication

Traditional feature extraction methods are used in manually identifying key patterns such as edges, textures, and shapes, enabling effective image representation for classification and object detection. It is essential to conduct before DL, as it provides structured feature descriptors for ML models. Most exploited methods are,

**Hu moments:** Hu Moments are a set of seven invariant image moments introduced by Ming-Kuei Hu [19] in 1962. It is widely used in pattern recognition and image analysis. These moments are derived from central moments and remain invariant to translation, scale, and rotation.

**Zernika moments:** Zernike Moments (ZMs) [20] are a set of orthogonal moments based on Zernike polynomials, widely used for image shape representation and pattern recognition. These moments are invariant to rotation and provide a robust feature descriptor.

**Harlick features:** Haralick features [21], introduced by Robert M. Haralick in 1973, are a set of statistical texture descriptors derived from the Gray Level Co-occurrence Matrix (GLCM). These features quantify spatial relationships between pixel intensities in an image, making them useful in texture classification and medical imaging.

**LDP features:** Local Directional Pattern (LDP) [22] is a texture descriptor that enhances feature extraction by considering edge responses in multiple directions. Unlike Local Binary Patterns (LBP), which use pixel intensity differences, LDP relies on edge detection, making it more robust to noise and illumination changes.

**Color Histogram:** A Color Histogram [23] is a statistical representation of the color distribution of an image. It quantifies the frequency of pixel intensities for each color channel, making it invariant to translation and rotation. Color histograms are widely used in image retrieval, classification, and object recognition.

### 3.3 DL feature extraction

Deep Learning (DL) has revolutionized feature extraction by automating hierarchical feature learning through Convolutional Neural Networks (CNNs). CNNs extract low-level edges, textures, and high-level patterns, enabling robust representation learning. Various architectures balance accuracy, efficiency, and depth, including VGG, ResNet, MobileNet, Inception, DenseNet, and Xception. VGG networks [24] (VGG16, VGG19) use stacked 3×3 convolutions to capture spatial hierarchies, making them effective for transfer learning. ResNet [25] introduced skip connections to overcome vanishing gradients, stabilizing training for deeper networks (ResNet50, ResNet101). MobileNet [26] optimizes efficiency with depth-wise separable convolutions, ideal for mobile applications. Inception [27] enhances multi-scale feature extraction using parallel convolutional pathways. InceptionV3 improves efficiency with factorized convolutions, while Inception-ResNetV2 integrates residual connections for higher accuracy. DenseNet [28] promotes feature reuse by passing feature maps across layers, improving efficiency (DenseNet121, DenseNet169). Xception [15] extends Inception by replacing standard convolutions with depth-wise separable convolutions, enhancing feature separation and computational efficiency. It surpasses InceptionV3 with fewer parameters, making it ideal for high-accuracy applications. These architectures drive advancements in classification, detection, and segmentation, optimizing feature extraction in modern DL models.

### 3.4 Proposed Methodology

In this study, we have proposed a classification framework that incorporates segmentation to filter out noise and extract the essential features. Afterward, we have introduced a modified feature extraction method that combines traditional feature extraction techniques, including Hu moments, Zernike moments, Haralick features, Local Binary Patterns (LBP) features, and color histograms, along with MobileNetV2 model-based deep feature extraction. These extracted features are then concatenated and used as input for a machine-learning classification model to categorize diabetic retinopathy into five distinct classes, as illustrated in Figure 2. The proposed method follows a structured model flow consisting of five key steps to classify diabetic retinopathy using a combination of segmentation, traditional feature extraction, deep feature extraction, and ML classification. Initially, retinal images are acquired and resized to a uniform dimension of (224 × 224 × 3) to ensure consistency across the dataset. These images then undergo segmentation, where relevant regions are extracted to highlight critical features necessary for effective diabetic retinopathy classification.

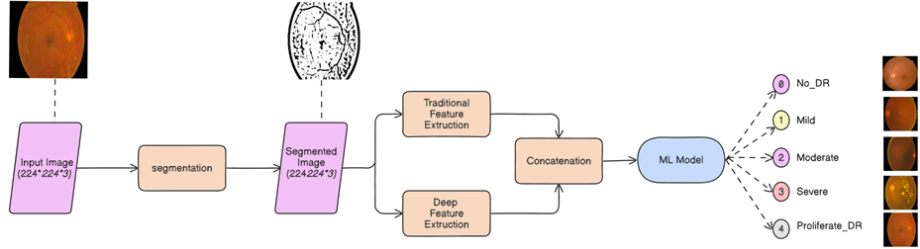

**Fig. 2.** Framework of the proposed model and image classes

Once the segmentation is complete, two distinct feature extraction techniques are applied: traditional feature extraction, which involves computing handcrafted statistical and structural features, and deep feature extraction, leveraging a DL model to obtain high-level representations of the retinal images. After extracting features through the traditional feature extraction processes and deep feature extraction process, a feature fusion process is carried out by concatenating the traditional and deep features into a comprehensive feature set, which enhances the model's capability of effectively distinguishing different stages of diabetic retinopathy. Finally, the fused feature set is fed into an ML classification model that categorizes the retinal images into five distinct classes: No_DR (0), Mild (1), Moderate (2), Severe (3), and Proliferative_DR (4), ensuring an accurate and robust classification framework for diabetic retinopathy detection.

## 4　Result Analysis and Discussion

Various DL models, including ResNet50, VGG-16, VGG-19, ResNet101, MobileNetV2, MobileNet, InceptionV3, InceptionResNetV2, DenseNet169, DenseNet121, and XceptionNet, were employed for feature extraction from the test dataset consisting of 3,622 images. The dataset was partitioned into 80% for model training (2,898) and 20% for model validation (724). Following feature extraction, different classification models—K-Nearest Neighbors (KNN), Support Vector Machine (SVM), Random Forest (RF), AdaBoost Classifier, and XGBoost Classifier—were utilized to classify diseases based on the extracted features. The experiments for this study were conducted using PyTorch and TensorFlow libraries within Python. Training for all classification models was performed in a Google Colab environment, utilizing three CUDA 12.4-enabled GPUs. The Diabetic Retinopathy dataset served as the benchmark.

The accuracy, sensitivity, and specificity of the classification results are summarized in Table 1. Our Model demonstrates superior recall, precision, F1 score, and Kappa values, with SVM achieving the best overall balance across all metrics. The Kappa score for SVM (0.49) and RF (0.47) is significantly higher than those observed in other models, indicating a stronger agreement between predictions and actual classifications.

**Table 1.** Table captions should be placed above the tables.

| Feature Extraction Backbone | Classifier | Accuracy | Recall | Precision | Kappa | F1 |
|---|---|---|---|---|---|---|
| ResNet50 | KNN | 52.42% | 0.52 | 0.47 | 0.22 | 0.34 |
| | SVM | 57.87% | 0.58 | 0.46 | 0.29 | 0.34 |
| | RF | 56.36% | 0.56 | 0.52 | 0.27 | 0.47 |
| | AdaBoost Classifier | 54.24% | 0.54 | 0.45 | 0.25 | 0.43 |
| | XGB Classifier | 54.24% | 0.54 | 0.5 | 0.23 | 0.47 |
| VGG-16 | KNN | 45.15% | 0.45 | 0.38 | 0.05 | 0.44 |
| | SVM | 51.51% | 0.52 | 0.0 | 0.27 | 0.34 |
| | RF | 49.39% | 0.49 | 0.32 | -0.01 | 0.5 |
| | AdaBoost Classifier | 51.21% | 0.51 | 0.35 | 0.02 | 0.49 |
| | XGB Classifier | 49.09% | 0.49 | 0.46 | 0.16 | 0.52 |
| VGG-19 | KNN | 46.06% | 0.46 | 0.42 | 0.07 | 0.39 |
| | SVM | 49.09% | 0.49 | 0.35 | 0.0 | 0.34 |
| | RF | 49.7% | 0.5 | 0.37 | 0.07 | 0.39 |
| | AdaBoost Classifier | 49.09% | 0.49 | 0.35 | 0.05 | 0.35 |
| | XGB Classifier | 47.27% | 0.47 | 0.38 | 0.1 | 0.47 |
| esNet101 | KNN | 43.94% | 0.44 | 0.33 | 0.02 | 0.51 |
| | SVM | 49.39% | 0.49 | 0.24 | 0.0 | 0.57 |
| | RF | 50.0% | 0.5 | 0.37 | 0.11 | 0.54 |
| | AdaBoost Classifier | 51.51% | 0.52 | 0.49 | 0.14 | 0.53 |
| | XGB Classifier | 44.85% | 0.45 | 0.36 | 0.07 | 0.59 |
| MobileNetV2 | KNN | 48.18% | 0.48 | 0.42 | 0.16 | 0.51 |
| | SVM | 58.48% | 0.58 | 0.43 | 0.28 | 0.52 |
| | RF | 60.0% | 0.6 | 0.51 | 0.33 | 0.54 |
| | AdaBoost Classifier | 58.79% | 0.59 | 0.53 | 0.32 | 0.47 |
| | XGB Classifier | 53.33% | 0.53 | 0.49 | 0.26 | 0.52 |
| MobileNet | KNN | 47.58% | 0.48 | 0.38 | 0.11 | 0.47 |
| | SVM | 50.61% | 0.51 | 0.26 | 0.0 | 0.36 |
| | RF | 56.67% | 0.57 | 0.54 | 0.26 | 0.47 |
| | AdaBoost Classifier | 49.09% | 0.49 | 0.43 | 0.08 | 0.16 |
| | XGB Classifier | 53.64% | 0.54 | 0.47 | 0.23 | 0.53 |
| Feature Extraction Backbone | Classifier | Accuracy | Recall | Precision | Kappa | F1 |
| InceptionV3 | KNN | 59.69% | 0.6 | 0.54 | 0.36 | 0.54 |
| | SVM | 60.0% | 0.6 | 0.48 | 0.35 | 0.54 |
| | RF | 61.21% | 0.61 | 0.56 | 0.38 | 0.59 |
| | AdaBoost Classifier | 61.21% | 0.61 | 0.64 | 0.36 | 0.56 |
| | XGB Classifier | 59.39% | 0.59 | 0.55 | 0.36 | 0.62 |
| InceptionRes-NetV2 | KNN | 49.09% | 0.49 | 0.33 | -0.0 | 0.2 |
| | SVM | 50.61% | 0.51 | 0.26 | 0.0 | 0.34 |
| | RF | 50.61% | 0.51 | 0.38 | 0.01 | 0.34 |
| | AdaBoost Classifier | 50.61% | 0.51 | 0.26 | 0.0 | 0.34 |
| | XGB Classifier | 47.88% | 0.48 | 0.36 | 0.0 | 0.34 |
| DenseNet169 | KNN | 47.27% | 0.47 | 0.35 | 0.02 | 0.32 |
| | SVM | 47.88% | 0.48 | 0.38 | 0.02 | 0.33 |
| | RF | 48.18% | 0.48 | 0.48 | 0.03 | 0.33 |
| | AdaBoost Classifier | 48.18% | 0.48 | 0.48 | 0.03 | 0.33 |
| | XGB Classifier | 46.97% | 0.47 | 0.45 | 0.02 | 0.33 |
| DenseNet121 | KNN | 46.06% | 0.46 | 0.22 | -0.01 | 0.3 |
| | SVM | 46.97% | 0.47 | 0.22 | 0.0 | 0.3 |
| | RF | 45.76% | 0.46 | 0.22 | -0.01 | 0.29 |
| | AdaBoost Classifier | 45.76% | 0.46 | 0.22 | -0.02 | 0.29 |
| | XGB Classifier | 43.64% | 0.44 | 0.22 | -0.04 | 0.29 |
| XceptionNet | KNN | 48.79% | 0.49 | 0.4 | 0.12 | 0.42 |
| | SVM | 46.97% | 0.47 | 0.22 | 0.0 | 0.3 |
| | RF | 46.36% | 0.46 | 0.37 | 0.0 | 0.31 |
| | AdaBoost Classifier | 46.97% | 0.47 | 0.22 | 0.0 | 0.3 |
| | XGB Classifier | 50.61% | 0.51 | 0.48 | 0.17 | 0.46 |
| Our Models | KNN | 63.44% | 0.59 | 0.63 | 0.43 | 0.60 |
| | SVM | 71.20% | 0.68 | 0.67 | 0.49 | 0.59 |
| | RF | 67.94% | 0.67 | 0.67 | 0.47 | 0.61 |
| | AdaBoost Classifier | 64.53% | 0.58 | 0.64 | 0.43 | 0.60 |
| | XGB Classifier | 60.00% | 0.51 | 0.58 | 0.37 | 0.58 |

This suggests that the proposed hybrid feature extraction approach enhances model robustness and generalization, improving classification performance. Comparing these results with the individual backbones, it is evident that the combination of traditional and deep feature extraction methods provides a substantial boost in accuracy and reliability. In deep feature extraction, the Random Forest (RF) classifier achieved the highest accuracy with several backbone models, including VGG-19 (49.7%), MobileNetV2 (60.0%), MobileNet (56.67%), InceptionV3 (61.21%), InceptionResNetV2 (50.61%), and DenseNet169 (48.18%). However, ResNet101 performed best with the AdaBoost classifier, achieving an accuracy of 51.51%, while XceptionNet yielded its highest accuracy of 50.61% with the XGBoost classifier. For ResNet50, VGG-16, DenseNet121, and Our Model, the SVM classifier proved to be the most effective, achieving 57.87%, 51.51%, 46.97%, and 71.0% accuracy, respectively. Figure 3 depicts heatmaps provided for visual comparison of different feature extraction backbones (ResNet50, VGG-16, MobileNet, InceptionV3, etc.) combined with various classifiers (KNN,

SVM, RF, AdaBoost, XGB) based on five key performance metrics: Accuracy, Recall, Precision, Kappa Score, F1-Score.

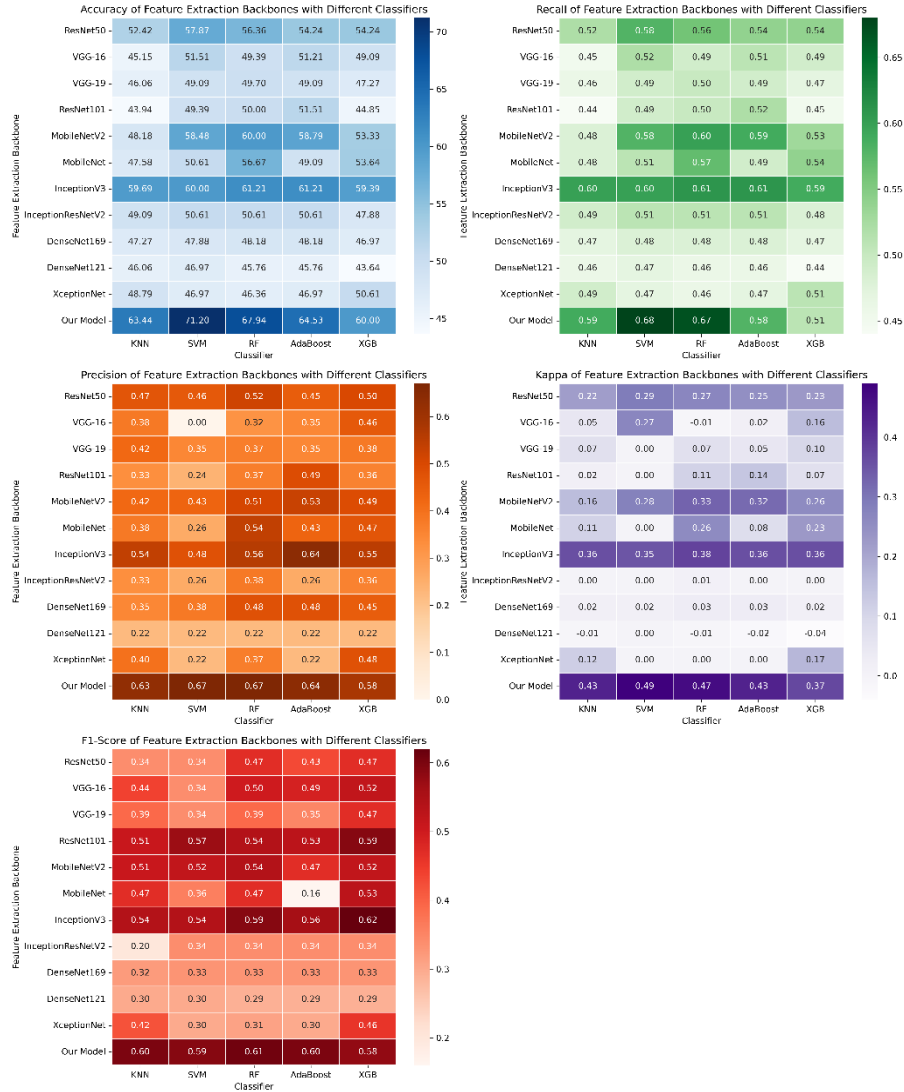

**Fig. 3.** Framework of the proposed model and image classes

Each cell in the heatmap represents a numerical value corresponding to a specific model-classifier combination, with a color gradient to make performance differences visually clear. Notably, our model outperformed all pre-trained architectures, showing the highest accuracy, recall, precision, and F1-score. InceptionV3 and MobileNetV2 were the best among pre-trained models, offering a balance of accuracy and sensitivity. DenseNet121 and InceptionResNetV2 had the weakest performance across all metrics, making them unsuitable for DR detection. Random Forest (RF) and SVM consistently performed well across multiple feature extraction backbones, proving to be the best classifiers. This superior performance highlights the effectiveness of the proposed hybrid approach, making it a highly suitable model for Diabetic Retinopathy classification.

## 5       Conclusions

In this study, we developed a modified classification architecture that integrates segmentation, traditional, and DL feature extraction techniques with ML-based classification. The proposed method demonstrates superior performance in classifying diabetic retinopathy when compared with the other state-of-the-art deep-learning feature extraction models. Within the dataset, segmentation aids in noise removal and enhances the extraction of essential features. The combination of traditional and DL feature extraction techniques significantly improves classification accuracy on the diabetic retinopathy dataset compared to existing state-of-the-art algorithms. For future research, we aim to address the high memory consumption of these models by exploring more efficient alternatives, such as quantum object detection models tested on more complex datasets.